\pdfoutput=1

\documentclass[11pt]{article}

\usepackage{acl}

\usepackage{times}
\usepackage{latexsym}

\usepackage[T1]{fontenc}

\usepackage[utf8]{inputenc}

\usepackage{microtype}
\usepackage{graphicx} 
\usepackage{subcaption} 
\usepackage{paralist} 
\usepackage{enumitem} 
\newcommand*{\blue}{\textcolor{black}}
\newcommand{\bl}[1]{\textcolor{black}{#1}}
\title{Utterance Emotion Dynamics in Children's Poems:\\ Emotional Changes Across Age}

\author{Daniela Teodorescu, Alona Fyshe\\
  University of Alberta\\
  \texttt{\small{\{dteodore,alona\}@ualberta.ca}} \\\And
  Saif M. Mohammad\\
  National Research Council Canada \\
  \texttt{\small saif.mohammad@nrc-cnrc.gc.ca} \\}

\begin{document}
\maketitle
\begin{abstract}

    \bl{Emerging} psychopathology studies \bl{are showing} that 
   \bl{patterns of} changes 
    in emotional state
\bl{ --- \textit{emotion dynamics} ---}
    are associated with 
     overall well-being and mental health. 
    \bl{More recently,} there has been some work in tracking emotion dynamics through one's utterances, 
    allowing for data to be collected on a larger scale 
    across time and people.
However, several questions about how 
\bl{emotion dynamics change with age, especially in children, 
and when determined through children's writing,} 
remain unanswered.
 In this work, we 
     use both a lexicon and a machine learning based approach to quantify characteristics of 
     \bl{emotion dynamics determined from poems written by children of various ages.
     We show that both approaches point to similar trends:
     consistent increasing intensities for some emotions (e.g., anger, fear, joy, sadness, arousal, and dominance) 
     with age
     and a consistent decreasing valence with age. 
     We also find increasing emotional variability, rise rates (i.e., emotional reactivity), and recovery rates (i.e., emotional regulation) with age.
These results act as a useful baselines for further research in how patterns of emotions expressed by children change with age, and their association with mental health.
          } 

\end{abstract}

\section{Introduction}
 \setitemize[0]{leftmargin=*}
 \setenumerate[0]{leftmargin=*}

Emotions play a key role in overall well-being
\citep{KUPPENS201722,Houben_2015,depressedYouth,sperry}. People's emotional states are constantly changing in response to internal and external events, and the way in which we regulate emotions
\citep{regulation2, regulation}.
Patterns of emotion change over time 
have been shown to be related to
general well-being and psychopathology (the scientific study of mental illness and disorders)
\citep{Houben_2015, sperry,internalExternal,Sheppes}, 
academic success \citep{success}, and 
social interactions in children \citep{Sosa}. 

Several psychopathology studies have introduced metrics  
to quantify and understand the trajectories and patterns in emotions across time \citep{KUPPENS201722}. These metrics are referred to as \textit{Emotion Dynamics} and include features of the emotional episode (e.g., duration) and of the emotional trajectory (e.g., emotional variability, co-variation, inertia) \citep{KUPPENS201722}. In psychology, emotion dynamics have usually been captured through self-report surveys over periods of time (e.g., five times a day for ten days).
However, obtaining such self-reports is arduous work; limiting the amount of data collected. Further, self-reports are prone to a number of biases (e.g., social pressures to be perceived as being happy).

Inspired by the emotion dynamics work in psychology, \citet{movieED} recently introduced the idea that patterns of emotion change can also be explored in the utterances of an individual, which can reflect their inner emotion dynamics.
They refer to this as \textit{utterance emotion dynamics (UED).}
They generate emotion arcs from streams of text (e.g., sentences in a story, tweets over time, etc.),
which are in turn used to determine various UED metrics.\footnote{An emotion arc is a series of time step--emotion value pairs that acts as a digital representation of how one's emotions change over time. There are several works in NLP that capture emotion arcs from streams of text (e.g., sentences in a story, tweets over time, etc.) \citep{mohammad-2011-upon,MOHAMMAD2012730,emotionarcs,teodorescu2022frustratingly,teodorescu2023generating}.}
Different UED metrics capture different aspects of emotion change (e.g., variability, rate of change, etc.).

\citet{teodorescu2022frustratingly} performed experiments on 36 diverse datasets to show that the quality of emotion arcs generated using emotion lexicons is comparable to those generated from machine learning (ML) methods.
The lexicon approach is able to 
perform well
through the power of aggregating information (e.g., 50--100 instances per bin).
\bl{Moreover,} the lexicon approach obtains high performance even when using translations of an English lexicon into low-resource languages, such as indigenous African languages \cite{teodorescu2023generating}. 
Emotion lexicons have the benefit of interpretability, accessibility, and efficiency compared to ML models. 
Thus, we \bl{primarily} used a lexicon-based approach \bl{in our experiments. However, we also show that the use of ML models points to the same trends as discovered by the lexicon approach.}\footnote{We did not find any poem datasets annotated for emotions that could be used to train an ML model; so we fine-tuned a pretrained ML model on emotion annotated tweets.}




UED metrics, calculated from emotion arcs, can be computed for a \textit{speaker} over time (e.g., main character in a narrative, tweets of a user over time), for multiple speakers at a time 
(e.g., treating all users in a geographic region as a \textit{speaker} for whom we can compute
UED),
or at an \textit{instance} level (e.g., independent posts where we compute UED metrics per post). 
While emotion dynamics have been studied 
in psychology for 
the past decades, UED was proposed only recently and has been applied 
to only a small number of domains (literature and tweets).
Important questions such as 
\textit{how do UED metrics change over development from toddlers to young adults?} and
\textit{how do the metrics change across one's adult life?}, remain unanswered.

 Generally, children's writing is a less studied domain in NLP, and there is limited data available. Also, research regarding children has guidelines and regulations in place to protect this vulnerable section of society \citep{poki}. 
 Yet, careful and responsible work such as the work done on the Child Language Data Exchange System (CHILDES) \cite{macwhinney2014childes} for understanding child language acquisition can be tremendously influential.
 Similarly, applying UED metrics to children's writing will allow us to infer the emotional states of children across age. 
 Such work provides important information for 
 psychologists and child development specialists,
 as emotion dynamics have been shown to underlie well-being, psychopathology, and success.

Poetry is a domain of growing interest in NLP (e.g., poem generation \citep{van-de-cruys-2020-automatic, 
goncalo-oliveira-2017-survey}). Poems are written to evoke emotions \citep{poemEmotional,JOHNSONLAIRD} and a medium through which emotions are expressed \citep{Whissell,belfi2018individual}.
The intersection of poems and children's writing
is an unexplored area which has the potential to unlock patterns in emotion word 
usage
by children as they age.

In this paper we contribute to the knowledge of emotion change 
over time
as children age
by studying poems written by children.
Our goal is to apply existing NLP techniques to study emotion change in childrens'
writing
rather than developing a novel algorithm for better detecting emotion.
We investigate the following questions:



\begin{compactitem}
\item How do the \textit{average} emotions vary across grades? How does this compare for discrete emotions (e.g., anger, fear, joy, and sadness) and emotion dimensions (e.g., valence, arousal, and dominance)? 
\item How \textit{variable} are emotion changes?
\end{compactitem}

\noindent These first two questions help us set crucial metrics in UED, building on work by \citet{poki}.
Next, to better understand patterns in emotion changes
we look at:
\begin{compactitem}
\item How does the rate at which children reach peak emotional states (\textit{rise rate}) change with age? Rise rate is analogous to emotional reactivity, which is associated with well-being. 
\item How does the rate at which children recover from peak emotional states back to 
steady state
(\textit{recovery rate}) change with age? Recovery rate plays a role in emotion regulation, 
which is also associated with well-being. 
\item How do \textit{utterance emotion dynamics} compare for adults vs.\@ children? 
\end{compactitem}

Answers to these questions provide baseline metrics for emotion change in children's poems across age.
In order to answer these questions, we use a dataset of $\sim61K$ poems written by children \citep{poki}
to calculate various UED metrics and examine how they vary across age.
The scores for the metrics and the analysis will act as useful baselines for further research
on emotion dynamics in children's writing, and their implications on mental health and well-being.

\section{Related Work}
Below we review related work on emotion dynamics and its ties
to well-being, the UED framework, and previous work on children's texts.

\subsection{Emotion Dynamics}
\label{sec:relatdWork_ed}
The \textit{emotion dynamics} framework studies change in emotion over time as it is
key to the study of emotions and overall well-being \citep{Houben_2015,depressedYouth}. Emotion dynamics metrics include \textit{emotion intensity} and \textit{emotion variability}. Emotion intensity is the average emotion over time. Whereas emotion variability is how much emotion changes from the average, often expressed as the standard deviation. These metrics have been used in various contexts in psychology to better understand well-being, often through self-reports or ecological momentary assessments.  


\blue{The relationship between various metrics in emotion dynamics and well-being have been the topic of numerous psychology studies.}
Higher positive and negative affect variability have been shown to be associated with lower psychological well-being and more mental health symptoms in
youth \citep{silk,Roekel} and adults \citep{Houben_2015}. 
\citet{Houben_2015} showed that emotion variability has significant correlation with numerous psychological well-being categories:
positive correlation with negative emotionality (e.g., negative affect and neuroticism), depression (e.g., depressive symptoms, depressive diagnosis), anxiety, borderline personality disorder, etc.
On the other hand, emotional variability is negatively correlated with self-esteem, quality of life, and other signs of high psychological well-being \citep{Houben_2015}.

\begin{table}[t]
\centering
{\small
\begin{tabular}
{lrr}
\hline
\multicolumn{1}{l}{\textbf{Dataset}} &
\multicolumn{1}{l}{\textbf{\# of Poems}} &
\multicolumn{1}{l}{\textbf{\#Words per Poem}} \\ \hline
\textbf{PoKi} & 61,330 & 14.3 \\
\hspace{3mm}Grade 1 & 900 & 37.3\\
\hspace{3mm}Grade 2 & 3,174 & 32.1\\
\hspace{3mm}Grade 3 & 6,712 & 35.2\\
\hspace{3mm}Grade 4 & 10,899 & 39.3\\
\hspace{3mm}Grade 5 & 11,479 & 44.5\\
\hspace{3mm}Grade 6 & 11,011 & 49.6\\
\hspace{3mm}Grade 7 & 7,831 & 59.7\\
\hspace{3mm}Grade 8 & 4,546 & 67.6\\
\hspace{3mm}Grade 9 & 1,284 & 91.5 \\
\hspace{3mm}Grade 10 & 1,171 & 91.8\\
\hspace{3mm}Grade 11 & 667 & 103.0\\
\hspace{3mm}Grade 12 & 1,656& 97.2\\[3pt] 
\textbf{FPP} &  50 & 181.02\\[3pt] 
     \hline
\end{tabular}
}
\caption{
Number of poems and the average lengths of poems in \textit{PoKi} (by grade) and in \textit{FPP}.
}
\vspace*{-6mm}
 \label{tab:poems}
\end{table}


Similarly, a vast number of studies explored the relationship between emotional regulation and reactivity with overall well-being. \citet{dysregulation} showed that mood and anxiety disorders are a result of emotion dysregulation of negative emotions, along with lacking positive emotions. Likewise, emotion dysregulation is thought of as the core of anxiety disorders \citep{MENNIN2007284,Carthy2010}. Children with anxiety disorders had higher negative emotion reactivity, and were less successful at implementing emotion regulation strategies \citep{Carthy2010}. 

\subsection{Utterance Emotion Dynamics}
As work in psychology measures emotion dynamics through self-report measures, emotion dynamics can also be determined from text using 
NLP 
techniques such as sentiment analysis. 

The UED framework \citep{movieED} tracks emotions dynamics in 
utterances, inspired by metrics in psychology. Such metrics include:
\begin{compactitem}
\item \textit{Home base}: The steady (most common) state where one is on average in emotional space. 
\item \textit{Variability}: How much one's emotional state changes with time. 
\item \textit{Rise Rate}: The rate at which one reaches peak emotional intensity, i.e., emotional reactivity. 
\item \textit{Recovery Rate}: The rate at which one recovers from peak emotional intensity to home base, i.e., emotional regulation. 
\end{compactitem}

This framework 
was used to study emotion arcs of movie characters \citep{movieED}, and to analyze 
emotional patterns across geographic regions through Twitter data 
\citep{vishnubhotla-mohammad-2022-tusc}.
\citet{depemodyn} studied the association between depression severity and emotion dynamics metrics such as variability 
on Facebook and Twitter. It was found that 
 increased negative emotional variability was an indicator for lower depression severity on Twitter.

 

\subsection{Children's Writing}
Few work studies children's writing due to the limited data available. One of the most commonly known datasets is the Child Language Data Exchange System (CHILDES) \cite{macwhinney2014childes} and in French, E-CALM \cite{doquet}. These datasets are limited in that they contain parent-child dialogue for children approximately age one to seven, and have limited quantities of text. 

Very few works look at emotions in children's writing. \citet{Manabe} performed sentiment analysis on narratives written by youth for mental illness detection, as self-disclosure is not the norm in some cultures. Participants wrote an imaginative story and answered a questionnaire on their tendencies toward psychological distress. It was found that youth who had higher tendencies toward psychological distress used significantly more positive words, and therefore had higher valence. 
In this work, we study the patterns of emotion word changes in poems written by children.  

\section{Poem Datasets}
\label{sec:poems}

For our experiments, we used a dataset of poems written by children as well as a dataset of poems written by adults (as control). 
Table \ref{tab:poems} shows key statistics of each dataset.

\noindent \textbf{Poems Written by Children (PoKi):}
\label{sec:poki} 
\citet{poki} compiled and curated a dataset of close to 61 thousand poems written by children in grades one to twelve. The poems were published and publicly available on the Scholastic Corporation website.\footnote{\citet{poki} obtained permission to use these poems for research.} In the PoKi dataset each poem is released with the child's school grade (which can be used as a proxy for age) and first name. 

The average emotional patterns for emotion dimensions (valence, arousal, and dominance), along with discrete emotions (anger, fear, joy and sadness) were analysed across grades. 
Additionally, these patterns were contrasted to those found in poems written by adults (data described 
below).
It was found that as children grow from early childhood into adolescence, valence decreases reaching a minimum at grade 11. Whereas arousal increases with age, aligning with how adults display emotions more visibly \citep{PMID:24821576,PMID:20809855}. Likewise, dominance increases with age. Consistently there was higher arousal in poems written by 
children 
with names commonly among males compared to those with names common among females. 
All intensities for anger, fear, joy and sadness increased across grades, with a particularly strong increase in sadness. 

\noindent \textbf{Poems Written by Adults (FPP):}
\citet{poki} also compiled and used poems written by adults which were published on the \textit{Famous Poets and Poems} website.\footnote{ 
We will refer to this dataset as \textit{FPP}. 
The poems are publicly available online and contain works by famous writers such as Edgar Allan Poe, and E.E. Cummings. 

\noindent \textbf{Preprocessing:} We preprocessed both poem datasets by removing extra whitespace, \blue{punctuation}, unescaping HTML (if any), tokenizing and lowercasing the text using the Twokenize\footnote{ 

\begin{figure*}
    \centering
    \begin{subfigure}[b]{0.49\textwidth}
        \centering
        \includegraphics[width=\textwidth]
        {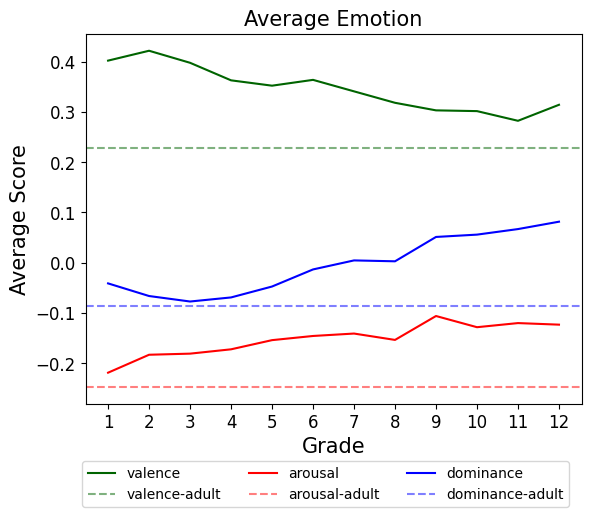}
        \caption{Valence, arousal and dominance}
        \label{fig:emo_mean}
    \end{subfigure}
    \hfill
    \begin{subfigure}[b]{0.49\textwidth}
        \centering
        \includegraphics[width=\textwidth]
        {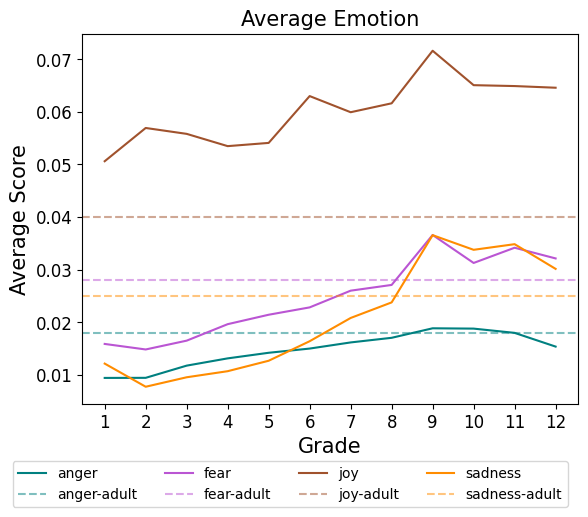}
        \caption{Anger, fear, joy, and sadness}
        \label{fig:emo_mean_discrete}
    \end{subfigure}
    \vspace*{-3mm}
    \caption{Average emotion across grades. The horizontal dashed lines represent values in poems written by adults.
    }
    \label{fig:emo_mean_2}
\end{figure*}

\section{Types of UED Metrics}
\bl{In the past, UED metrics have been calculated for the speaker or jointly for text from a set of speakers (meta-speaker).} 
We propose 
a third form of UED metrics not explored before --- \textit{instance} level UED metrics.
\bl{All three of these types of UED metrics are described below:}\\[-18pt]

\begin{itemize}
    \item \textbf{Speaker UED Metrics:} Here all available utterances by a speaker are placed in temporal order to form the text from which the UED metrics for the speaker is determined. For each metric, UED scores from multiple speakers can be averaged to determined the average UED score for that metric for the population.  In the past, speaker UED metrics have been determined for characters in movie dialogues \citep{movieED}, and for users on Twitter during the pandemic \citep{vishnubhotla-mohammad-2022-tusc}.\\[-18pt]
    \item \textbf{Meta-Speaker UED Metrics:} If one is interested in analyzing change of emotions in a discourse by multiple speakers, for example, analyzing changes in emotion patterns in a Reddit thread, then we can treat each discourse (e.g., each Reddit thread) as text produced by a meta-speaker. Here we arrange each of the utterances in each of the discourses (e.g., Reddit thread) in temporal order and determine the UED metrics for each discourse. UED metrics for all of the discourses can be averaged to determine the average UED metric scores for 
    a 
    set of discourses. In the past, discourse UED metrics have been determined for users from geographic regions, such as treating all users on Twitter in a country as a speaker \citep{vishnubhotla-mohammad-2022-tusc}.\\[-18pt]
     \item \textbf{Instance UED Metrics:} If one is interested in the change of emotions in 
     individual 
     pieces of 
      text (or instances) such as a novel, a poem, a tweet, or a blog post, 
      then we can simply apply the UED metrics to each instance.
      \bl{Such a metric is useful at individual instance level if the instance is long enough (otherwise the score for the metric is not a reliable on its own).
      However, even for smaller pieces of text, the UED scores from a large number of instances can give a reliable estimate of the distribution of these instance-level UED metrics. Such metrics can even be used to compare patterns of emotion change across different sets, where each set is composed of 
       (a) instances from many speakers and (b)
           instances that are temporally unordered (either because that information is not available or because we are not interested in temporal ordering of items within a set). Examples of instance-level UED use include:
           determining UEDs of presidential speeches,
           comparing average UEDs of stream of consciousness essays of different age groups, etc.}\\[-18pt]
\end{itemize}

\noindent In this work we are interested in children's poems (instances of poetic text) across age and not how each individual child has a different writing style. Therefore, we calculate UED metrics for each poem and average the scores for each grade.

\begin{figure*}
    \centering
    \begin{subfigure}[b]{0.49\textwidth}
        \centering
        \includegraphics[width=\textwidth]
        {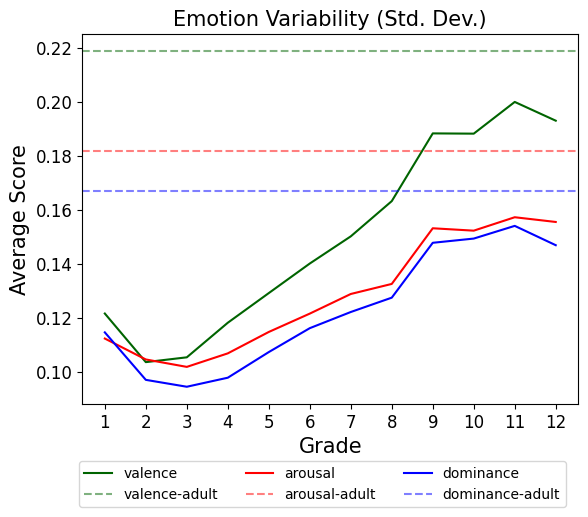}
        \caption{Valence, arousal and dominance}
        \label{fig:emo_std}
    \end{subfigure}
    \hfill
    \begin{subfigure}[b]{0.49\textwidth}
        \centering
        \includegraphics[width=\textwidth]
        {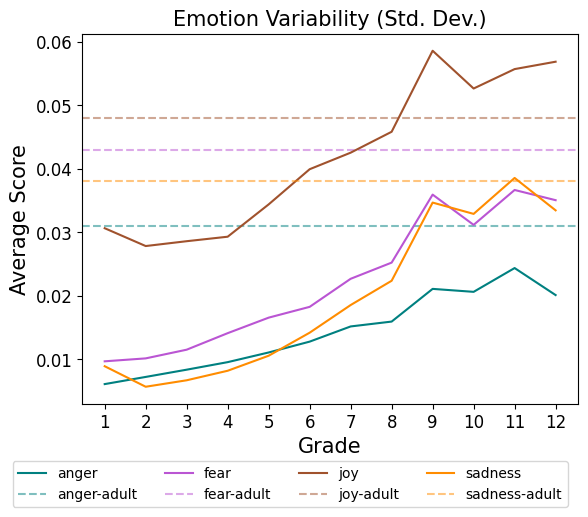}
        \caption{Anger, fear, joy, and sadness}
        \label{fig:emo_std_discrete}
    \end{subfigure}
      \vspace*{-3mm}
    \caption{Emotion variability across grades. The horizontal dashed lines represent values in poems written by adults.
    }
    \label{fig:emo_std_2}
\end{figure*}

\begin{table*}[!ht]
\centering
{\small
\begin{tabular}
{lrrrrrrrl}
\hline
\multicolumn{1}{l}{\textbf{UED Metric}} &
\multicolumn{1}{l}{\textbf{Valence}} &
\multicolumn{1}{l}{\textbf{Arousal}} &
\multicolumn{1}{l}{\textbf{Dominance}} &
\multicolumn{1}{l}{\textbf{Anger}} &
\multicolumn{1}{l}{\textbf{Fear}} &
\multicolumn{1}{l}{\textbf{Joy}} &
\multicolumn{1}{l}{\textbf{Sadness}} &
\multicolumn{1}{l}{\textbf{Psych. Construct}}\\ \hline

Average & 0.228 & -0.247 & -0.087 & 0.018 & 0.028 & 0.040 & 0.025 & Intensity \\

Variability & 0.219 & 0.182 & 0.167 & 0.031 & 0.043 & 0.048 & 0.038 & Emotional Variability\\


Rise Rate & 0.134 & 0.114 & 0.084 & 0.115 & 0.109 & 0.066 & 0.113 & Emotional Reactivity\\

Recovery Rate & 0.127 & 0.105 & 0.086 & 0.024 & 0.028 & 0.023 & 0.020 & Emotional Regulation\\ 

\hline
\end{tabular}
}
\vspace*{-1mm}
\caption{ The values for UED metrics in poems written by adults, and the corresponding construct in psychology.}
\vspace*{-3mm}
\label{tab:adult_ued}
\end{table*}

\section{Experiments}
\label{sec:experiments}
\bl{Our goal is to} analyze how patterns of emotion words change with age \bl{in children's poems}. In order to do so, we generate an emotion arc per poem and compute instance-level UED metrics.
Afterwards, we average the UED metrics per grade 
to compare results across age.
We use the Emotion Dynamics toolkit\footnote{https://github.com/Priya22/EmotionDynamics} to calculate UED metrics and our code for the experiments is available online.\footnote{https://github.com/dteodore/EmotionArcs}

We use text windows of size five \blue{(excluding words with a neutral emotion score)} and a step size of one to create an emotion arc per poem.
\blue{We only considered poems 
that included at least five emotion words\footnote{as per the NRC VAD lexicon}.}
For each research question we computed the corresponding metrics: average emotional state, emotional variability, 
rise rate and recovery rate.   
Analyzing average emotion and variability allows us to build foundational knowledge into changes in patterns of emotion words. We then look at
rise rate and recovery rate to further our understanding of children's emotion dynamics.

While older children (e.g., grade 10--12) tend to write on average longer poems than younger children (e.g., grade 1--3), these UED metrics are not affected by the length of the poems.\blue{\footnote{We show in Appendix \ref{appendix:length_metrics} that similar patterns in UED metrics hold when controlling for poem length across grades.}}
Other metrics calculate the number of displacements from home base or the length of displacements to peaks which are affected by 
poem length.
Additionally, because poems are shorter than text streams such as novels, the number of windows that can be created from a poem is limited, so metrics specific to 
\blue{emotional}
displacement 
are not computed 
since they 
are more suitable for longer texts.

Each metric is computed for both dimensional emotions (e.g., valence, arousal, dominance) and discrete emotions (e.g., anger, fear, joy and sadness). We used the NRC Valence, Arousal, and Dominance (VAD) Lexicon \citep{vad-acl2018} and the NRC Emotion Intensity Lexicon \citep{LREC18-AIL} for word-emotion scores.

In Section \ref{sec:results} we use the lexicon-based approach to generate emotion arcs.
We explain how the metrics are computed, contrast the trends across grades 
and 
compare the results to poems written by adults. 
We discuss the ties of these results with work in psychology and implications for emotional development.
In Section \ref{sec:ml_approach} we explore the same questions using an ML model for generating emotion arcs. We 
find similar trends across grade
\bl{with the ML approach as when using} 
the lexicon approach.

\subsection{Utterance Emotion Dynamics: PoKi}
\label{sec:results}

We begin with a question on how average emotion word score changes with grade--a question that \citet{poki}
already answered in their work. We replicated the experiment to make sure any differences in preprocessing the data or code development did not lead to different results. We 
then answer the other questions 
on how specifically do the trajectories of emotion change across grade differ,
which have not been addressed yet.
Likewise, we compute the UED metrics on the poems written by adults. We show the results in Table \ref{tab:adult_ued} and contrast them 
to PoKi below. 

\begin{figure*}
    \centering
    \begin{subfigure}[t]{0.49\textwidth}
    \centering
    \includegraphics[width=\textwidth]
    {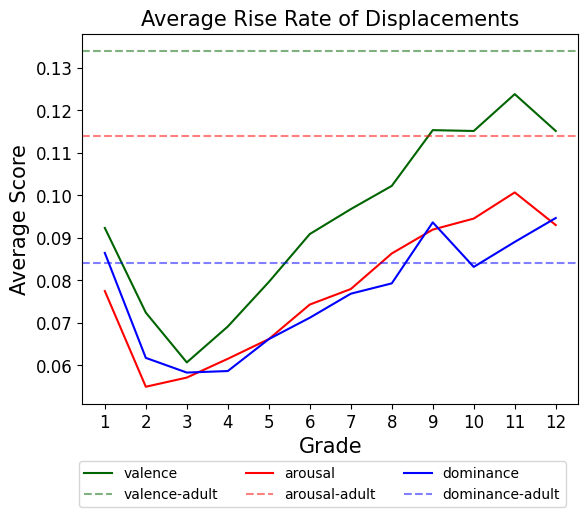}
    \caption{Valence, arousal and dominance}
     \label{fig:rise_rate}
    \end{subfigure}
    \hfill
    \begin{subfigure}[t]{0.49\textwidth}
    \centering
    \includegraphics[width=\textwidth]
    {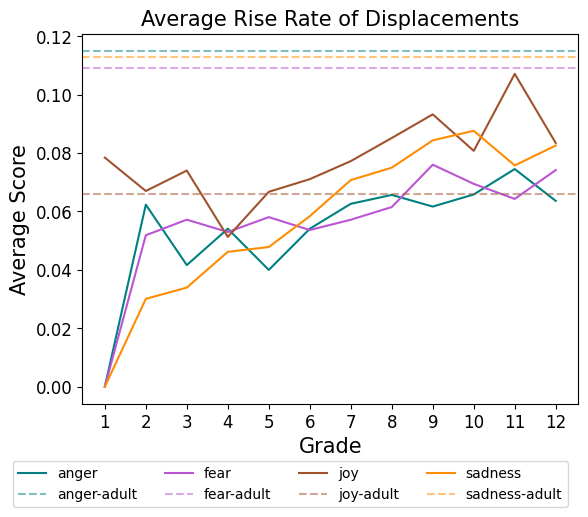}
    \caption{Anger, fear, joy and sadness}
     \label{fig:rise_rate_discrete}
    \end{subfigure}
    \caption{Rise rate in poems across grades. The horizontal dashed lines represent values in poems written by adults.
    }
    \label{fig:rise_rate_2}
   \vspace*{-3mm}
\end{figure*}

\subsubsection{How does the average emotion expressed change across age?}

\noindent \textbf{Method:} An average emotion score is calculated per window in the poem using word-emotion scores from the lexicon, and then the average is computed across windows in a poem.

\noindent \textbf{Results:} Below we present results on both the valence, arousal, and dominance (VAD) dimensions
as well as for discrete emotion categories (Anger, Fear, Joy, Sadness).

\noindent \textit{PoKi VAD:} In Figure \ref{fig:emo_mean}, we show the average VAD emotions expressed across grade. Overall, we see a downward trend in valence from Grade 1 to Grade 12. This means that the poems written by younger children have, on average, more positive emotion words than those written by older children. There is a slight peak at grade 6, however a consistent downwards trend overall. Arousal and dominance similarly both trend upwards with age. There is a steeper increase for arousal and dominance at grade 9. This means that children are expressing more active and powerful emotions 
in 
poems as they age.

\noindent \textit{FPP VAD:}
The average valence of 0.228 
is notably lower than the valence across grades, where the lowest is reached by grade 11s at 
0.28. 
The average arousal at -0.247 and dominance at -0.087 are lower than those of children 
across all ages, and interestingly most similar to 
younger children.


\noindent \textit{PoKi Anger, Fear, Joy, Sadness:}
In Figure \ref{fig:emo_mean_discrete}, we see that the average 
discrete emotions 
all increase across grades. Anger, while increasing from grade 1 to 9, has a downward trend from grade 10 to 12. All emotions tend to have a peak around grade 9 \blue{and plateau afterwards}.

\noindent \textit{FPP Anger, Fear, Joy, Sadness:}
Anger, fear and sadness tends to match to those of older children around grade 8 to 9. 
Children from grade 9 to 12 reach even higher values than adults for fear and sadness.
On the other hand, joy tends to remain below those of children across all age, and has the most similar values to younger children at 0.04.



\noindent \textbf{Discussion:}
These findings align with those by \citet{poki} which similarly computed the mean emotion in poems across grade. 
Numerous works in psychology have found similar trends through self-report studies for valence \citep{FROST2015132,larson,simmons,Weinstein}, and arousal \citep{Carstensen,Gunnar,Somerville}. Likewise, as sadness increased with age, \citet{Holsen} have shown that teenagers are more likely to experience a negative and depressed mood. This trend matters because we are seeing similar trends in the emotion words 
used by children when writing poems
as those in psychology self-reports, although they were not told to explicitly talk about how they are feeling. This work further contributes to the current findings on emotional development in children. 

\subsubsection{How variable are emotions across age?}

\noindent \textbf{Method:} 
Variability is computed as the standard deviation of emotion values for windows in a poem. 

\noindent \textbf{Results:}\\ 
\textit{PoKi VAD}: In Figure \ref{fig:emo_std}, variability for valence, arousal, and dominance all trend upward with age; stabilizing in grades 11 and 12.

\noindent \textit{FPP VAD}: 
For all three emotions variability was most similar to those of older children, reaching slightly above 
grades 10--12. 

\noindent \textit{PoKi Anger, Fear, Joy, Sadness}:
In Figure \ref{fig:emo_std_discrete}, we see that variability for all emotions trend upwards from grade 1 to 9, and start to level out around grade 10 to 12. Anger, fear, and sadness all have a peak at grade 9 and grade 11. Joy has an especially pronounced peak at grade 9. 

\noindent \textit{FPP Anger, Fear, Joy, Sadness}:
Variability in anger, fear and sadness is higher for adults than those expressed by children across all grades, and is most similar to older children around grade 11. 
Likewise, variability for joy in adults is more similar to older children, however around grade 8.

\begin{figure*}
\vspace*{-6mm}
    \centering
    \begin{subfigure}[t]{0.49\textwidth}
    \centering
    \includegraphics[width=\textwidth]
    {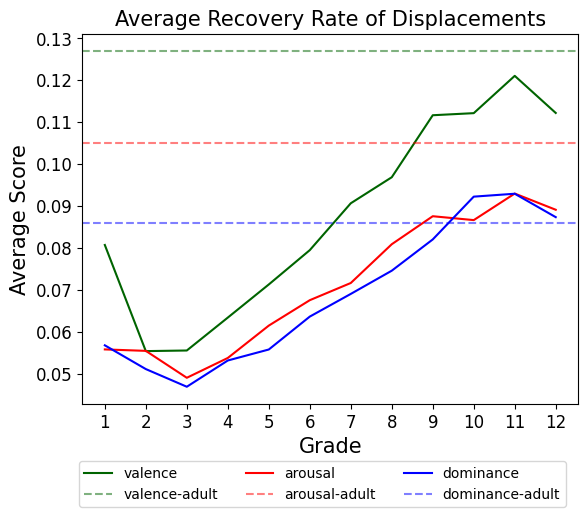}
    \caption{Valence, arousal and dominance}
    \label{fig:recovery_rate}
    \end{subfigure}
    \hfill
    \begin{subfigure}[t]{0.495\textwidth}
    \centering
    \includegraphics[width=\textwidth]
    {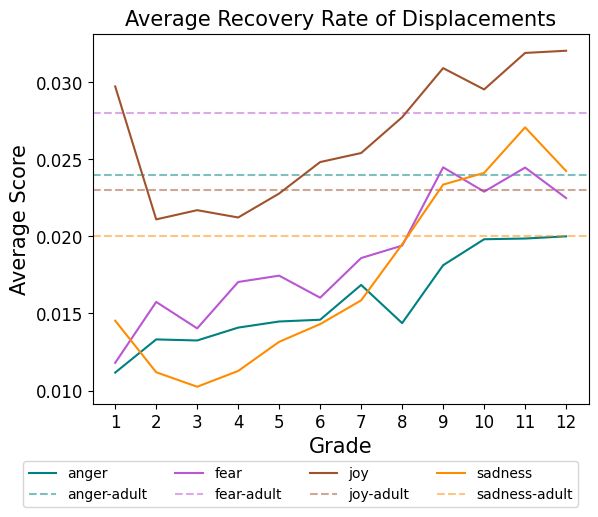}
    \caption{Anger, fear, joy and sadness}
    \label{fig:recovery_rate_discrete}
    \end{subfigure}
     \vspace*{-3mm}
    \caption{Recovery rate in poems across grades. The horizontal dashed lines represent values for  poems by adults.
    }
    \label{fig:recovery_rate_2}
    \vspace*{-3mm}
\end{figure*}



\noindent \textbf{Discussion:}
The overall trend of increasing emotional variability with age, followed by 
stabilizing
supports findings in psychology. 
\citet{larson} found that emotional variability increased over early adolescence and stabilized around mid-adolescence. 
Further, during adolescence important cognitive, social and psychical changes occur which are thought to increase emotional variability \citep{Buchanan,Arnett, STEINBERG200569}. 
\citet{Reitsema} found that sadness variability statistically increased with age. 
These trends are important as they support those found in psychology which are strongly associated with mental well-being \citep{Reitsema}.

\subsubsection{At what rate do emotions change from home to peak state?} 

\noindent \textbf{Method:} The average rise rate is calculated as the average of the rise rate for 
windows
in a poem. The rise rate 
is \textit{peak distance} (how far away the peak is from the home base) divided by the number of words during the rise period. The rise rate 
disregards
the direction of the peak.

\noindent \textbf{Results:} 

\noindent \textit{PoKi VAD}: In Figure \ref{fig:rise_rate}, we see that rise rate increases for all three emotions across grade, and plateaus around grade 10 to 12. The rise rate is comparably higher for valence, followed by arousal and then dominance. 

\noindent \textit{FPP VAD}:
The rise rate for valence and arousal in adults is higher than those across all
 grades,
 and is most similar to older students in grade 11. The rise rate for dominance in adults also matches those of older children, however 
 starting at
 grade 8
 (with grade 9, 11 and 12 having a higher rate than adults). 


\noindent \textit{PoKi Anger, Fear, Joy, Sadness}:
In Figure \ref{fig:rise_rate_discrete}, the rise rate for the discrete emotions all increase with grade. 
Joy has a small dip around grade 4, and then increases matching the average rise rate of anger, fear, and sadness which all started at slightly lower values in grade 2. 
We note that at grade 1 we could not compute the average rise rate for anger, fear, and sadness as the poems had too few displacements (the number of poems which had displacements was less than our pre-chosen threshold of 5).

\noindent \textit{FPP Anger, Fear, Joy, Sadness}:
The rise rate for anger, fear, and sadness in adult poems is higher than those expressed in children across all ages, with most similar values to older children. However, the rise rate for joy is lower 
and corresponds with younger children around grade 2 to 5. 



\noindent \textbf{Discussion:}
Rise rate is seen as analogous to reactivity in psychology, which has been found to increase during adolescence \citep{somerville2016emotional}. Our findings 
support 
these trends. As mentioned in \ref{sec:relatdWork_ed}, emotional reactivity is at the core of anxiety and attention disorder, impacting overall well-being.

\begin{figure*}
\vspace*{-6mm}
    \centering
    \begin{subfigure}[b]{0.4\textwidth}
    \centering
    \includegraphics[width=\textwidth]{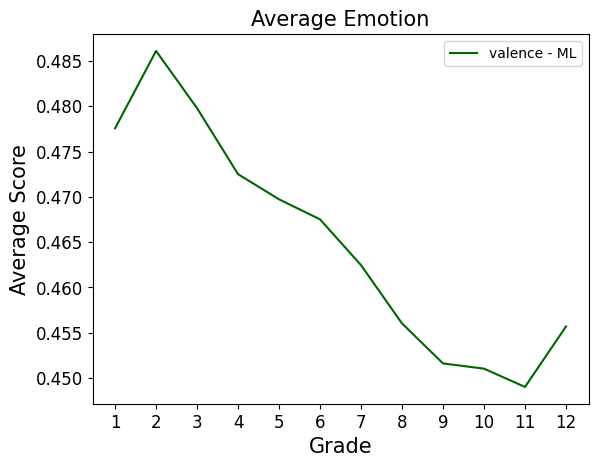}
    \label{fig:emo_mean_ml}
    \end{subfigure}
    \hfill
    \begin{subfigure}[b]{0.4\textwidth}
    \centering
    \includegraphics[width=\textwidth]
    {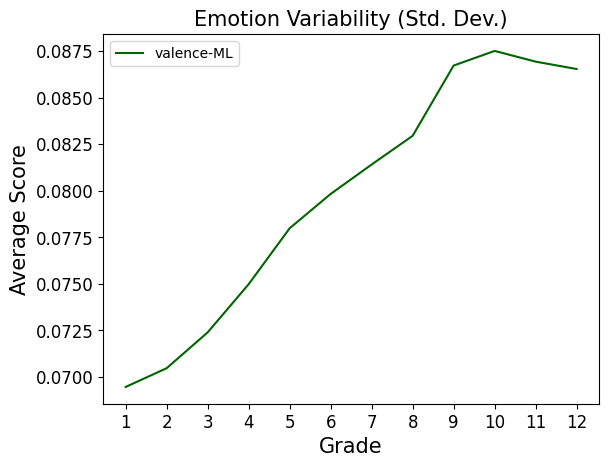}
    \label{fig:emo_std_ml}
    \end{subfigure}
    \vspace*{-9mm}
    \caption{Average emotion and emotion variability for valence using the ML \textit{n-gram} approach on the PoKi dataset.
    }
    \label{fig:ml_results}
    \vspace*{-3mm}
\end{figure*}

\subsubsection{At what rate do emotions recover?} 

\noindent \textbf{Method:} Recovery rate is computed similarly 
to
rise rate, however 
divides peak distance by the number of words during the recovery period.
Recovery rate does not distinguish between 
peak direction.

\noindent \textbf{Results}:

\noindent \textit{PoKi VAD}:
Figure \ref{fig:recovery_rate} we see the recovery rate increases for all three emotions with age and plateaus around grade 10 to 12. While the valence recovery rate has a larger magnitude
than the other emotions,
all rates trend upwards. Recovery rate can be thought of \textit{emotion regulation}, 
indicating
that older children are able to return to their home base emotional states after a peak
more quickly than younger children.

\noindent \textit{FPP VAD}:
The recovery rate of adults for valence and arousal corresponds most closely with 
older children (e.g., grades 9--12), however is higher than 
across all grades. 
The recovery rate of dominance is similar to grade 9 students, and slightly below those of older children.


\noindent \textit{PoKi Anger, Fear, Joy, Sadness}:
In Figure \ref{fig:recovery_rate_discrete}, 
we
similarly
see increasing average recovery rates for all 4 emotions across age. The magnitude 
of joy's recovery rate  
is considerably higher than for the other 3 emotions. 

\noindent \textit{FPP Anger, Fear, Joy, Sadness}:
The recovery rate for fear and anger is above those across all ages, and is most similar to older children in grade 9--12. On the other hand, the recovery rate for joy and sadness matches those of younger children, around grade 5 for joy and 8 for sadness.  



\noindent \textbf{Discussion:}
Recovery rate, which is analogous to emotion regulation, has been studied extensively in psychology. \citet{Zeman} detail the progression of emotional regulation from infancy to adolescence, in which an increase in emotion regulation occurs alongside developments in strategies and motivations. Not only does emotion regulation have ties with well-being, it also plays a role in academic success of children \citep{success} and adults \citep{Phillips}.

\subsection{Utterance Emotion Dynamics - ML Approach: PoKi}
\label{sec:ml_approach}
\blue{To perform a comprehensive analysis and ensure the trends in emotion change are consistent regardless of the emotion labelling method used, we also performed experiments using a ML model. 
Previously, individual words were emotion labelled using a lexicon.
Now, we use a \textit{n-gram} approach where a ML model assigns emotion scores to windows of text in the poem of length \textit{n}. 
We are not trying to determine which of these two approaches is \textit{better} at computing UED metrics as this would be challenging - there are no existing annotated datasets for emotion arcs, or UED metrics. Rather, we are supporting the trends found by the word-level lexicon approach, with those found by ML models as they are commonly used on downstream tasks (e.g., sentiment analysis) and are known for their strong performance. If the ML approach did not perform well, we would not expect any trends in UED metrics to appear. 
}

\noindent \blue{\textbf{Datasets:} We use the same poem datasets 
as 
in Section \ref{sec:poems}, creating n-gram windows of length 5. The only difference is that words not found in the emotion lexicon or neutral words can be included. We choose this approach as ML models are trained on sequential text.
}

\noindent \blue{\textbf{Experiments:} We fine-tuned a RoBERTa \citep{liu2019roberta} base model for fine-grain sentiment analysis using the SemEval 2018 Task 1 dataset \citep{SemEval2018Task1}. This means that we were able to predict an emotion score between 0 and 1. Details on the model training are in Appendix \ref{sec:ml_details}.
After emotion labelling text windows, 
we performed similar experiments as in Section \ref{sec:experiments}: compute the UED metrics per poem and take the average per grade for each metric.
}




\noindent \blue{\textbf{Results:} Overall we found similar trends as with the \textit{word-level} lexicon approach. We note that a direct comparison between the lexicon and ML approach can not be made as they are using different units of measurement (e.g., windows 
 contain either sequential words found in the lexicon or natural sequences of words).
 We can simply compare the trends in emotion change rather than the magnitude of change or the values themselves.
We discuss the results for valence below
(Figure \ref{fig:ml_results} and Appendix \ref{appendix:ml_results}).
We also show the results for the discrete emotions in Appendix \ref{appendix:ml_results}, as 
the trends were similar 
 to the lexicon approach.
}

\noindent \blue{\textbf{Average Emotion}: As grade increases, we see a similar downwards trend and a stabilization at grades 10--12.
}

\noindent \blue{\textbf{Emotional Variability}: Older children tend to show increased variability.
}

\noindent \blue{\textbf{Rise Rate and Recovery Rate}:
With age, children are writing with increased rates of emotional reactivity and also emotion regulation.
}

\blue{Overall, these results show that there are patterns of emotion change in childrens' poems with age, and trends found using the lexicon 
approach are also replicated using ML models.}

\section{Conclusion}
We explored four utterance emotion dynamics metrics (average, variability, rise rate, and recovery rate), 
and seven emotions (three dimensional and four discrete) on poems written by children and adults. 
We found that the patterns of emotion changes in poetry by children supported previous results and findings in the psychology literature (e.g., increased variability, rise rate, and recovery rates with age). 

As future work, we would like to examine poetry by adults more in-depth, 
such as 
how do patterns of emotion change look 
for
experts vs.\@ 
novices?
And how do UED compare across geographic regions, and time periods.

\section*{Limitations}
A limitation of this work is that the poems written by adults are by experienced writers who are often known for their poetry. These poems may therefore not be representative of poems written by adults in general, and could affect the patterns and trends in emotion words we see. As future work we would like to expand the collection of poems written by adults to include those written by novices as well.

\section*{Ethics Statement}
\noindent Our research interest is to study emotions at an aggregate/group level. This has applications in emotional development psychology and in public health (e.g., overall well-being and mental health). However, emotions are complex, private, and central to an individual's experience. Additionally, each individual expresses emotion differently through language, which results in large amounts of variation. Therefore, several ethical considerations should be accounted for when performing any textual analysis of emotions \cite{Mohammad22AER,mohammad2020practicaleacl}.
The ones we would particularly like to highlight are listed below:
\begin{compactitem}
    \item Our work on studying emotion word usage should not be construed as detecting how people feel; rather, we draw inferences on the emotions that are conveyed by users via the language that they use. 
    \item The language used in an utterance may convey information about the emotional state (or perceived emotional state) of the speaker, listener, or someone mentioned in the utterance. However, it is not sufficient for accurately determining any of their momentary emotional states. Deciphering true momentary emotional state of an individual requires extra-linguistic context and world knowledge.
    Even then, one can be easily mistaken.
    \item The inferences we draw in this paper are based on aggregate trends across large populations. We do not draw conclusions about specific individuals or momentary emotional states.
\end{compactitem}

\section*{Acknowledgements}
We thank Will Hipson for collecting the PoKi poems dataset and for setting the groundwork in this direction of research. We also thank Krishnapriya Vishnubhotla for the toolkit she created to help compute UED metrics.

\bibliography{anthology,custom}

\begin{thebibliography}{56}
\expandafter\ifx\csname natexlab\endcsname\relax\def\natexlab#1{#1}\fi

\bibitem[{Arnett(1999)}]{Arnett}
JJ~Arnett. 1999.
\newblock \href {https://doi.org/10.1037//0003-066x.54.5.317} {Adolescent storm
  and stress, reconsidered}.
\newblock \emph{The American psychologist}, 54(5):317—326.

\bibitem[{Belfi et~al.(2018)Belfi, Vessel, and Starr}]{belfi2018individual}
Amy~M Belfi, Edward~A Vessel, and G~Gabrielle Starr. 2018.
\newblock Individual ratings of vividness predict aesthetic appeal in poetry.
\newblock \emph{Psychology of Aesthetics, Creativity, and the Arts}, 12(3):341.

\bibitem[{Buchanan et~al.(1992)Buchanan, Eccles, and Becker}]{Buchanan}
CM~Buchanan, JS~Eccles, and JB~Becker. 1992.
\newblock \href {https://doi.org/10.1037/0033-2909.111.1.62} {Are adolescents
  the victims of raging hormones: evidence for activational effects of hormones
  on moods and behavior at adolescence}.
\newblock \emph{Psychological bulletin}, 111(1):62—107.

\bibitem[{Carstensen et~al.(2000)Carstensen, Pasupathi, Mayr, and
  Nesselroade}]{Carstensen}
LL~Carstensen, M~Pasupathi, U~Mayr, and JR~Nesselroade. 2000.
\newblock \href {https://doi.org/10.1037/0022-3514.79.4.644} {Emotional
  experience in everyday life across the adult life span}.
\newblock \emph{Journal of personality and social psychology}, 79(4):644—655.

\bibitem[{Carthy et~al.(2010)Carthy, Horesh, Apter, and Gross}]{Carthy2010}
Tal Carthy, Netta Horesh, Alan Apter, and James~J. Gross. 2010.
\newblock \href {https://doi.org/10.1007/s10862-009-9167-8} {Patterns of
  emotional reactivity and regulation in children with anxiety disorders}.
\newblock \emph{Journal of Psychopathology and Behavioral Assessment},
  32(1):23--36.
\newblock This study was supported by a research fund of the Adler Research
  Center in Tel-Aviv University. The authors would like to thank the Anxiety
  Disorders Clinic in ‘Schneider’s Children Medical Center of Israel’ for
  support and collaboration. Special thanks to Ronit Jossifoff, Maya Ferber,
  Yael Tadmor and Hilit Pritsch for their important contribution to the
  recruitment and examination of the participants.

\bibitem[{Doquet(2013)}]{doquet}
Claire Doquet. 2013.
\newblock \href {https://hal.science/hal-01236152} {{Ancrages th{\'e}oriques de
  l'analyse g{\'e}n{\'e}tique des textes d'{\'e}l{\`e}ves}}.
\newblock In Catherine~Bore et~Eduardo~Calil, editor, \emph{{L'Ecole,
  l'{\'e}criture et la cr{\'e}ation. Etudes fran{\c c}aises et
  br{\'e}siliennes.}}, Sciences du langage - Carrefour et points de vue, pages
  33--53. {Academia Bruylant}.

\bibitem[{Dreyfuss et~al.(2014)Dreyfuss, Caudle, Drysdale, Johnston, Cohen,
  Somerville, Galván, Tottenham, Hare, and Casey}]{PMID:24821576}
Michael Dreyfuss, Kristina Caudle, Andrew~T Drysdale, Natalie~E Johnston,
  Alexandra~O Cohen, Leah~H Somerville, Adriana Galván, Nim Tottenham, Todd~A
  Hare, and BJ~Casey. 2014.
\newblock \href {https://doi.org/10.1159/000357755} {Teens impulsively react
  rather than retreat from threat}.
\newblock \emph{Developmental neuroscience}, 36(3-4):220—227.

\bibitem[{Frost et~al.(2015)Frost, Hoyt, Chung, and Adam}]{FROST2015132}
Allison Frost, Lindsay~T. Hoyt, Alissa~Levy Chung, and Emma~K. Adam. 2015.
\newblock \href
  {https://doi.org/https://doi.org/10.1016/j.adolescence.2015.06.001} {Daily
  life with depressive symptoms: Gender differences in adolescents' everyday
  emotional experiences}.
\newblock \emph{Journal of Adolescence}, 43:132--141.

\bibitem[{Gon{\c{c}}alo~Oliveira(2017)}]{goncalo-oliveira-2017-survey}
Hugo Gon{\c{c}}alo~Oliveira. 2017.
\newblock \href {https://doi.org/10.18653/v1/W17-3502} {A survey on intelligent
  poetry generation: Languages, features, techniques, reutilisation and
  evaluation}.
\newblock In \emph{Proceedings of the 10th International Conference on Natural
  Language Generation}, pages 11--20, Santiago de Compostela, Spain.
  Association for Computational Linguistics.

\bibitem[{Graziano et~al.(2007)Graziano, Reavis, Keane, and Calkins}]{success}
Paulo~A. Graziano, Rachael~D. Reavis, Susan~P. Keane, and Susan~D. Calkins.
  2007.
\newblock \href {https://doi.org/https://doi.org/10.1016/j.jsp.2006.09.002}
  {The role of emotion regulation in children's early academic success}.
\newblock \emph{Journal of School Psychology}, 45(1):3--19.

\bibitem[{Gunnar et~al.(2009)Gunnar, Wewerka, Frenn, Long, and Griggs}]{Gunnar}
Megan~R Gunnar, Sandi Wewerka, Kristin Frenn, Jeffrey~D Long, and Christopher
  Griggs. 2009.
\newblock \href {https://doi.org/10.1017/s0954579409000054} {Developmental
  changes in hypothalamus-pituitary-adrenal activity over the transition to
  adolescence: normative changes and associations with puberty}.
\newblock \emph{Development and psychopathology}, 21(1):69—85.

\bibitem[{Hipson and Mohammad(2020)}]{poki}
Will Hipson and Saif~M. Mohammad. 2020.
\newblock \href {https://aclanthology.org/2020.lrec-1.196} {{P}o{K}i: A large
  dataset of poems by children}.
\newblock In \emph{Proceedings of the Twelfth Language Resources and Evaluation
  Conference}, pages 1578--1589, Marseille, France. European Language Resources
  Association.

\bibitem[{Hipson and Mohammad(2021)}]{movieED}
Will~E. Hipson and Saif~M. Mohammad. 2021.
\newblock \href {https://doi.org/10.1371/journal.pone.0256153} {Emotion
  dynamics in movie dialogues}.
\newblock \emph{PLOS ONE}, 16(9):1--19.

\bibitem[{Hofmann et~al.(2012)Hofmann, Sawyer, Fang, and
  Asnaani}]{dysregulation}
Stefan~G. Hofmann, Alice~T. Sawyer, Angela Fang, and Anu Asnaani. 2012.
\newblock \href {https://doi.org/https://doi.org/10.1002/da.21888} {Emotion
  dysregulation model of mood and anxiety disorders}.
\newblock \emph{Depression and Anxiety}, 29(5):409--416.

\bibitem[{Holsen et~al.(2000)Holsen, Kraft, and Vitterso}]{Holsen}
Ingrid Holsen, Pal Kraft, and Joar Vitterso. 2000.
\newblock \href
  {https://login.ezproxy.library.ualberta.ca/login?url=https://www.proquest.com/scholarly-journals/stability-depressed-mood-adolescence-results-6/docview/204521468/se-2}
  {Stability in depressed mood in adolescence: Results from a 6-year
  longitudinal panel study}.
\newblock \emph{Journal of Youth and Adolescence}, 29(1):61--78.
\newblock Copyright - Copyright Plenum Publishing Corporation Feb 2000; Last
  updated - 2023-02-07; CODEN - JYADA6.

\bibitem[{Houben et~al.(2015)Houben, Noortgate, and Kuppens}]{Houben_2015}
Marlies Houben, Wim Van~Den Noortgate, and Peter Kuppens. 2015.
\newblock \href {https://doi.org/10.1037/a0038822} {The relation between
  short-term emotion dynamics and psychological well-being: A meta-analysis.}
\newblock \emph{Psychological Bulletin}, 141(4):901--930.

\bibitem[{Johnson-Laird and Oatley(2022)}]{JOHNSONLAIRD}
Philip~N. Johnson-Laird and Keith Oatley. 2022.
\newblock \href {https://doi.org/https://doi.org/10.1016/j.actpsy.2022.103506}
  {How poetry evokes emotions}.
\newblock \emph{Acta Psychologica}, 224:103506.

\bibitem[{Kuppens and Verduyn(2017)}]{KUPPENS201722}
Peter Kuppens and Philippe Verduyn. 2017.
\newblock \href {https://doi.org/https://doi.org/10.1016/j.copsyc.2017.06.004}
  {Emotion dynamics}.
\newblock \emph{Current Opinion in Psychology}, 17:22--26.
\newblock Emotion.

\bibitem[{Larson et~al.(2002)Larson, Moneta, Richards, and Wilson}]{larson}
Reed~W. Larson, Giovanni Moneta, Maryse~H. Richards, and Suzanne Wilson. 2002.
\newblock \href {https://doi.org/https://doi.org/10.1111/1467-8624.00464}
  {Continuity, stability, and change in daily emotional experience across
  adolescence}.
\newblock \emph{Child Development}, 73(4):1151--1165.

\bibitem[{Liu et~al.(2019)Liu, Ott, Goyal, Du, Joshi, Chen, Levy, Lewis,
  Zettlemoyer, and Stoyanov}]{liu2019roberta}
Yinhan Liu, Myle Ott, Naman Goyal, Jingfei Du, Mandar Joshi, Danqi Chen, Omer
  Levy, Mike Lewis, Luke Zettlemoyer, and Veselin Stoyanov. 2019.
\newblock Roberta: A robustly optimized bert pretraining approach.
\newblock \emph{arXiv preprint arXiv:1907.11692}.

\bibitem[{MacWhinney(2014)}]{macwhinney2014childes}
Brian MacWhinney. 2014.
\newblock \emph{The CHILDES project: Tools for analyzing talk, Volume II: The
  database}.
\newblock Psychology Press.

\bibitem[{Manabe et~al.(2021)Manabe, Liew, Yada, Wakamiya, and
  Aramaki}]{Manabe}
Masae Manabe, Kongmeng Liew, Shuntaro Yada, Shoko Wakamiya, and Eiji Aramaki.
  2021.
\newblock \href {https://doi.org/10.2196/29500} {Estimation of psychological
  distress in japanese youth through narrative writing: Text-based stylometric
  and sentiment analyses}.
\newblock \emph{JMIR Form Res}, 5(8):e29500.

\bibitem[{McRae et~al.(2012)McRae, Gross, Weber, Robertson, Sokol-Hessner, Ray,
  Gabrieli, and Ochsner}]{regulation}
Kateri McRae, James~J. Gross, Jochen Weber, Elaine~R. Robertson, Peter
  Sokol-Hessner, Rebecca~D. Ray, John~D.E. Gabrieli, and Kevin~N. Ochsner.
  2012.
\newblock \href {https://doi.org/10.1093/scan/nsr093} {{The development of
  emotion regulation: an fMRI study of cognitive reappraisal in children,
  adolescents and young adults}}.
\newblock \emph{Social Cognitive and Affective Neuroscience}, 7(1):11--22.

\bibitem[{Mennin et~al.(2007)Mennin, Holaway, Fresco, Moore, and
  Heimberg}]{MENNIN2007284}
Douglas~S. Mennin, Robert~M. Holaway, David~M. Fresco, Michael~T. Moore, and
  Richard~G. Heimberg. 2007.
\newblock \href {https://doi.org/https://doi.org/10.1016/j.beth.2006.09.001}
  {Delineating components of emotion and its dysregulation in anxiety and mood
  psychopathology}.
\newblock \emph{Behavior Therapy}, 38(3):284--302.

\bibitem[{Mohammad(2011)}]{mohammad-2011-upon}
Saif Mohammad. 2011.
\newblock \href {https://aclanthology.org/W11-1514} {From once upon a time to
  happily ever after: Tracking emotions in novels and fairy tales}.
\newblock In \emph{Proceedings of the 5th {ACL}-{HLT} Workshop on Language
  Technology for Cultural Heritage, Social Sciences, and Humanities}, pages
  105--114, Portland, OR, USA. Association for Computational Linguistics.

\bibitem[{Mohammad(2023)}]{mohammad2020practicaleacl}
Saif Mohammad. 2023.
\newblock \href {https://aclanthology.org/2023.findings-eacl.136} {Best
  practices in the creation and use of emotion lexicons}.
\newblock In \emph{Findings of the Association for Computational Linguistics:
  EACL 2023}, pages 1825--1836, Dubrovnik, Croatia. Association for
  Computational Linguistics.

\bibitem[{Mohammad(2012)}]{MOHAMMAD2012730}
Saif~M. Mohammad. 2012.
\newblock \href {https://doi.org/https://doi.org/10.1016/j.dss.2012.05.030}
  {From once upon a time to happily ever after: Tracking emotions in mail and
  books}.
\newblock \emph{Decision Support Systems}, 53(4):730--741.
\newblock 1) Computational Approaches to Subjectivity and Sentiment Analysis 2)
  Service Science in Information Systems Research : Special Issue on PACIS
  2010.

\bibitem[{Mohammad(2018{\natexlab{a}})}]{vad-acl2018}
Saif~M. Mohammad. 2018{\natexlab{a}}.
\newblock Obtaining reliable human ratings of valence, arousal, and dominance
  for 20,000 english words.
\newblock In \emph{Proceedings of The Annual Conference of the Association for
  Computational Linguistics (ACL)}, Melbourne, Australia.

\bibitem[{Mohammad(2018{\natexlab{b}})}]{LREC18-AIL}
Saif~M. Mohammad. 2018{\natexlab{b}}.
\newblock Word affect intensities.
\newblock In \emph{Proceedings of the 11th Edition of the Language Resources
  and Evaluation Conference (LREC-2018)}, Miyazaki, Japan.

\bibitem[{Mohammad(2022)}]{Mohammad22AER}
Saif~M. Mohammad. 2022.
\newblock Ethics sheet for automatic emotion recognition and sentiment
  analysis.
\newblock \emph{To Appear in Computational Linguistics}.

\bibitem[{Mohammad et~al.(2018)Mohammad, Bravo-Marquez, Salameh, and
  Kiritchenko}]{SemEval2018Task1}
Saif~M. Mohammad, Felipe Bravo-Marquez, Mohammad Salameh, and Svetlana
  Kiritchenko. 2018.
\newblock Semeval-2018 {T}ask 1: {A}ffect in tweets.
\newblock In \emph{Proceedings of International Workshop on Semantic Evaluation
  (SemEval-2018)}, New Orleans, LA, USA.

\bibitem[{Phillips et~al.(2002)Phillips, Phillips, Bull, Adams, and
  Fraser}]{Phillips}
Louise~H Phillips, Louise~H Phillips, Rebecca Bull, Ewan Adams, and Lisa
  Fraser. 2002.
\newblock \href {https://doi.org/10.1037/1528-3542.2.1.12} {Positive mood and
  executive function: evidence from stroop and fluency tasks}.
\newblock \emph{Emotion (Washington, D.C.)}, 2(1):12—22.

\bibitem[{Reagan et~al.(2016)Reagan, Mitchell, Kiley, Danforth, and
  Dodds}]{emotionarcs}
Andrew~J. Reagan, Lewis Mitchell, Dilan Kiley, Christopher~M. Danforth, and
  Peter~S. Dodds. 2016.
\newblock \href
  {https://www.proquest.com/scholarly-journals/emotional-arcs-stories-are-dominated-six-basic/docview/1865288690/se-2}
  {The emotional arcs of stories are dominated by six basic shapes}.
\newblock \emph{EPJ Data Science}, 5(1):1--12.
\newblock Copyright - EPJ Data Science is a copyright of Springer, 2016; Last
  updated - 2017-02-06.

\bibitem[{Reitsema et~al.(2022)Reitsema, Jeronimus, van Dijk, and
  de~Jonge}]{Reitsema}
Anne~M Reitsema, Bertus~F Jeronimus, Marijn van Dijk, and Peter de~Jonge. 2022.
\newblock \href {https://doi.org/10.1037/emo0000970} {Emotion dynamics in
  children and adolescents: A meta-analytic and descriptive review}.
\newblock \emph{Emotion (Washington, D.C.)}, 22(2):374—396.

\bibitem[{Scott et~al.(2020)Scott, Victor, Kaufman, Beeney, Byrd, Vine,
  Pilkonis, and Stepp}]{internalExternal}
Lori~N Scott, Sarah~E Victor, Erin~A Kaufman, Joseph~E Beeney, Amy~L Byrd, Vera
  Vine, Paul~A Pilkonis, and Stephanie~D Stepp. 2020.
\newblock \href {https://doi.org/10.1177/2167702619898802} {Affective dynamics
  across internalizing and externalizing dimensions of psychopathology}.
\newblock \emph{Clinical psychological science : a journal of the Association
  for Psychological Science}, 8(3):412—427.

\bibitem[{Seabrook et~al.(2018)Seabrook, Kern, Fulcher, and
  Rickard}]{depemodyn}
Elizabeth~M Seabrook, Margaret~L Kern, Ben~D Fulcher, and Nikki~S Rickard.
  2018.
\newblock \href {https://doi.org/10.2196/jmir.9267} {Predicting depression from
  language-based emotion dynamics: Longitudinal analysis of facebook and
  twitter status updates}.
\newblock \emph{J Med Internet Res}, 20(5):e168.

\bibitem[{Sheppes et~al.(2015)Sheppes, Suri, and Gross}]{Sheppes}
Gal Sheppes, Gaurav Suri, and James~J. Gross. 2015.
\newblock \href {https://doi.org/10.1146/annurev-clinpsy-032814-112739}
  {Emotion regulation and psychopathology}.
\newblock \emph{Annual Review of Clinical Psychology}, 11(1):379--405.
\newblock PMID: 25581242.

\bibitem[{Silk et~al.(2011)Silk, Forbes, Whalen, Jakubcak, Thompson, Ryan,
  Axelson, Birmaher, and Dahl}]{depressedYouth}
Jennifer~S. Silk, Erika~E. Forbes, Diana~J. Whalen, Jennifer~L. Jakubcak,
  Wesley~K. Thompson, Neal~D. Ryan, David~A. Axelson, Boris Birmaher, and
  Ronald~E. Dahl. 2011.
\newblock \href {https://doi.org/https://doi.org/10.1016/j.jecp.2010.10.007}
  {Daily emotional dynamics in depressed youth: A cell phone ecological
  momentary assessment study}.
\newblock \emph{Journal of Experimental Child Psychology}, 110(2):241--257.
\newblock Special Issue: Assessment of Emotion in Children and Adolescents.

\bibitem[{Silk et~al.(2003)Silk, Steinberg, and Morris}]{silk}
Jennifer~S Silk, Laurence Steinberg, and Amanda~Sheffield Morris. 2003.
\newblock \href {https://doi.org/10.1046/j.1467-8624.2003.00643.x}
  {Adolescents' emotion regulation in daily life: links to depressive symptoms
  and problem behavior}.
\newblock \emph{Child development}, 74(6):1869—1880.

\bibitem[{Simmons et~al.(1987)Simmons, Burgeson, Carlton-Ford, and
  Blyth}]{simmons}
RG~Simmons, R~Burgeson, S~Carlton-Ford, and DA~Blyth. 1987.
\newblock \href {https://doi.org/10.1111/j.1467-8624.1987.tb01453.x} {The
  impact of cumulative change in early adolescence}.
\newblock \emph{Child development}, 58(5):1220—1234.

\bibitem[{Somerville(2013)}]{Somerville}
Leah~H Somerville. 2013.
\newblock \href {https://doi.org/10.1177/0963721413476512} {Special issue on
  the teenage brain: Sensitivity to social evaluation}.
\newblock \emph{Current directions in psychological science}, 22(2):121—127.

\bibitem[{Somerville et~al.(2011)Somerville, Hare, and Casey}]{PMID:20809855}
Leah~H Somerville, Todd Hare, and BJ~Casey. 2011.
\newblock \href {https://doi.org/10.1162/jocn.2010.21572} {Frontostriatal
  maturation predicts cognitive control failure to appetitive cues in
  adolescents}.
\newblock \emph{Journal of cognitive neuroscience}, 23(9):2123—2134.

\bibitem[{Somerville(2016)}]{somerville2016emotional}
LH~Somerville. 2016.
\newblock Emotional development in adolescence.
\newblock \emph{Handbook of emotions}, pages 350--365.

\bibitem[{Sosa-Hernandez et~al.(2022)Sosa-Hernandez, Wilson, and
  Henderson}]{Sosa}
Linda Sosa-Hernandez, McLennon Wilson, and Heather~A Henderson. 2022.
\newblock \href {https://doi.org/10.1037/emo0001155} {Emotion dynamics among
  preadolescents getting to know each other: Dyadic associations with shyness}.
\newblock \emph{Emotion (Washington, D.C.)}.

\bibitem[{Sperry et~al.(2020)Sperry, Walsh, and Kwapil}]{sperry}
Sarah~H. Sperry, Molly~A. Walsh, and Thomas~R. Kwapil. 2020.
\newblock \href {https://doi.org/https://doi.org/10.1016/j.jad.2019.09.076}
  {Emotion dynamics concurrently and prospectively predict mood
  psychopathology}.
\newblock \emph{Journal of Affective Disorders}, 261:67--75.

\bibitem[{Steinberg(2005)}]{STEINBERG200569}
Laurence Steinberg. 2005.
\newblock \href {https://doi.org/https://doi.org/10.1016/j.tics.2004.12.005}
  {Cognitive and affective development in adolescence}.
\newblock \emph{Trends in Cognitive Sciences}, 9(2):69--74.

\bibitem[{Teodorescu and Mohammad(2022)}]{teodorescu2022frustratingly}
Daniela Teodorescu and Saif~M. Mohammad. 2022.
\newblock \href {http://arxiv.org/abs/2210.07381} {Frustratingly easy sentiment
  analysis of text streams: Generating high-quality emotion arcs using emotion
  lexicons}.

\bibitem[{Teodorescu and Mohammad(2023)}]{teodorescu2023generating}
Daniela Teodorescu and Saif~M. Mohammad. 2023.
\newblock \href {http://arxiv.org/abs/2306.02213} {Generating high-quality
  emotion arcs for low-resource languages using emotion lexicons}.

\bibitem[{Van~de Cruys(2020)}]{van-de-cruys-2020-automatic}
Tim Van~de Cruys. 2020.
\newblock \href {https://doi.org/10.18653/v1/2020.acl-main.223} {Automatic
  poetry generation from prosaic text}.
\newblock In \emph{Proceedings of the 58th Annual Meeting of the Association
  for Computational Linguistics}, pages 2471--2480, Online. Association for
  Computational Linguistics.

\bibitem[{van Roekel et~al.(2016)van Roekel, Bennik, Bastiaansen, Verhagen,
  Ormel, Engels, and Oldehinkel}]{Roekel}
Eeske van Roekel, Elise~C Bennik, Jojanneke~A Bastiaansen, Maaike Verhagen,
  Johan Ormel, Rutger C M~E Engels, and Albertine~J Oldehinkel. 2016.
\newblock \href {https://doi.org/10.1007/s10802-015-0090-z} {Depressive
  symptoms and the experience of pleasure in daily life: An exploration of
  associations in early and late adolescence}.
\newblock \emph{Journal of abnormal child psychology}, 44(5):999—1009.

\bibitem[{Vishnubhotla and Mohammad(2022)}]{vishnubhotla-mohammad-2022-tusc}
Krishnapriya Vishnubhotla and Saif~M. Mohammad. 2022.
\newblock \href {https://aclanthology.org/2022.lrec-1.442} {{Tweet Emotion
  Dynamics}: Emotion word usage in tweets from {US} and {C}anada}.
\newblock In \emph{Proceedings of the Thirteenth Language Resources and
  Evaluation Conference}, pages 4162--4176, Marseille, France. European
  Language Resources Association.

\bibitem[{Wassiliwizky et~al.(2017)Wassiliwizky, Koelsch, Wagner, Jacobsen, and
  Menninghaus}]{poemEmotional}
Eugen Wassiliwizky, Stefan Koelsch, Valentin Wagner, Thomas Jacobsen, and
  Winfried Menninghaus. 2017.
\newblock \href {https://doi.org/10.1093/scan/nsx069} {The emotional power of
  poetry: neural circuitry, psychophysiology and compositional principles}.
\newblock \emph{Social cognitive and affective neuroscience},
  12(8):1229—1240.

\bibitem[{Weinstein et~al.(2007)Weinstein, Mermelstein, Hankin, Hedeker, and
  Flay}]{Weinstein}
Sally~M. Weinstein, Robin~J. Mermelstein, Benjamin~L. Hankin, Donald Hedeker,
  and Brian~R. Flay. 2007.
\newblock \href
  {https://doi.org/https://doi.org/10.1111/j.1532-7795.2007.00536.x}
  {Longitudinal patterns of daily affect and global mood during adolescence}.
\newblock \emph{Journal of Research on Adolescence}, 17(3):587--600.

\bibitem[{Whissell(2004)}]{Whissell}
Cynthia Whissell. 2004.
\newblock \href {https://doi.org/10.2190/FWGA-M9DB-P9D4-11X6} {Poetic emotion
  and poetic style: The 100 poems most frequently included in anthologies and
  the work of emily dickinson}.
\newblock \emph{Empirical Studies of the Arts}, 22(1):55--75.

\bibitem[{Zeman et~al.(2006)Zeman, Cassano, Perry-Parrish, and Stegall}]{Zeman}
Janice Zeman, Michael Cassano, Carisa Perry-Parrish, and Sheri Stegall. 2006.
\newblock \href {https://doi.org/10.1097/00004703-200604000-00014} {Emotion
  regulation in children and adolescents}.
\newblock \emph{Journal of developmental and behavioral pediatrics : JDBP},
  27(2):155—168.

\bibitem[{Zimmermann and Iwanski(2014)}]{regulation2}
Peter Zimmermann and Alexandra Iwanski. 2014.
\newblock \href {https://doi.org/10.1177/0165025413515405} {Emotion regulation
  from early adolescence to emerging adulthood and middle adulthood: Age
  differences, gender differences, and emotion-specific developmental
  variations}.
\newblock \emph{International Journal of Behavioral Development},
  38(2):182--194.

\end{thebibliography}

\appendix

\section{Poem Length on UED Metrics}
\label{appendix:length_metrics}
As mentioned in Section \ref{sec:experiments}, 
certain UED metrics
which rely on distances (e.g., length of displacements to peaks) could be influenced by poem length.
Therefore, we selected metrics which are based on rates or averages. To verify these metrics are not impacted by the increasing poem lengths with age, we investigated if the same trends hold when controlling for the length of poems across grade. In Figure \ref{fig:emo_mean_length_2} we show the results for the average valence across grades for poems of length 10 to 20 words (not including stop words). 
As grade increases, we similarly see a decrease in valence. Similar trends occur with other metrics.

\begin{figure}[t]
    \centering
    \begin{subfigure}[b]{0.49\textwidth}
    \centering
    \includegraphics[width=\textwidth]{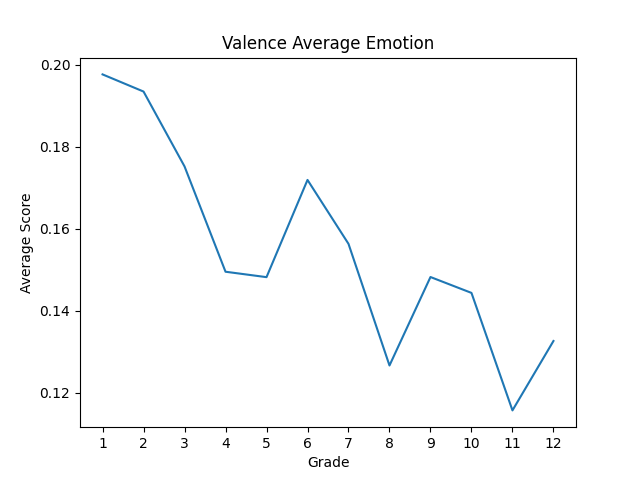}
     \label{fig:emo_mean_length}
    \end{subfigure}
    \caption{Average valence across grades for poems of length 10 to 20 words.}
    \label{fig:emo_mean_length_2}
\end{figure}

\section{Machine Learning Approach}
In Section \ref{sec:ml_details} we detail the model training process and in Section \ref{appendix:ml_results} we show the results on the PoKi dataset using a ML approach.

\subsection{Model Training}
\label{sec:ml_details}

\blue{
We fine-tuned the pretrained RoBERTa \citep{liu2019roberta} base model available on HuggingFace\footnote{\url{https://huggingface.co/roberta-base}}. For training we used the SemEval 2018 Task 1 dataset \citep{SemEval2018Task1} which contains tweets annotated with emotion scores for valence, anger, fear, joy and sadness.\footnote{We could not train models for arousal and dominance as there are no corresponding annotated datasets.} The dataset contains both fine-grain emotion scores (real-valued numbers between 0 and 1) and categorical labels (e.g., -1, 0, 1). We use the real-valued emotion scores to compute more fine-grained emotion arcs. More details on this dataset are available in Table \ref{tab:semeval_2018} and Table \ref{tab:data_splits}. 
}

\begin{table*}[h]
\centering
{\small
\begin{tabular}
{llllll}
\hline
\multicolumn{1}{l}{\textbf{Dataset}} &
\multicolumn{1}{l}{\textbf{Source}} &
\multicolumn{1}{l}{\textbf{Domain}} &
\multicolumn{1}{l}{\textbf{Dimension}} &
\multicolumn{1}{l}{\textbf{Label Type}} & \multicolumn{1}{l}{\textbf{\# Instances}}\\ \hline

SemEval 2018 (EI-Reg) & \citet{SemEval2018Task1} & tweets & anger, fear & continuous (0 to 1) & 3092, 3627,\\
     &  &   & joy, sadness   &  &  3011, 2095 \\[3pt]

SemEval 2018 (V-Reg) & \citet{SemEval2018Task1} & tweets & valence & continuous (0 to 1) & 2567\\\hline

\end{tabular}
}
\vspace*{-1mm}
\small
\caption{Dataset descriptive statistics. The No. of instances includes the train, dev, and test sets for the Sem-Eval 2018 Task 1 (EI-Reg and V-Reg).} 
 \label{tab:semeval_2018}
\end{table*}

\begin{table*}[h!]
\centering
{
\begin{tabular}
{lrrr}
\hline
\multicolumn{1}{l}{\textbf{Emotion}} &
\multicolumn{1}{l}{\textbf{Train}} &
\multicolumn{1}{l}{\textbf{Dev.}} &
\multicolumn{1}{l}{\textbf{Test}}\\ \hline
Valence & 1181 & 449 & 937\\
Anger & 1701 & 388 & 1002\\
Fear & 2252 & 389 & 986\\
Joy & 1616 & 290 & 1105\\
Sadness & 1533 & 397 & 975\\\hline
\end{tabular}
}
\vspace*{-1mm}
\caption{The number of tweets in each of the dataset splits for the SemEval 2018 Task 1.}  
 \label{tab:data_splits}
\end{table*}

\blue{
We used the Trainer pipeline from HuggingFace\footnote{\url{https://huggingface.co/docs/evaluate/main/en/transformers_integrations\#trainer}} to fine-tune the pretrained model. For the loss function we used mean-square loss.
}

\blue{
We tuned the following hyperparameters on the development set and selected the best model using mean-square error: learning rate (2e-5, 3e-5), number of epochs (5, 10, 20) and batch size (16, 32). Note that our aim here is not to overly fine-tune the model as we are applying it to a different domain (i.e., poems). The best parameters for each emotion model are shown in Table \ref{tab:hyperparameter_tune}.
After determining the \textit{best} model on the development set we apply it to windows of text in the PoKi poem dataset. 
}

\begin{table*}[h!]
\centering
{
\begin{tabular}
{lrrr}
\hline
\multicolumn{1}{l}{\textbf{Emotion}} &
\multicolumn{1}{l}{\textbf{Learning Rate}} &
\multicolumn{1}{l}{\textbf{No. Epochs}} &
\multicolumn{1}{l}{\textbf{Batch Size}}\\ \hline
Valence & 3e-05 & 32 & 5\\
Anger & 2e-05 & 32 & 10\\
Fear & 3e-05 & 32 & 10\\
Joy & 2e-05 & 16 & 5\\
Sadness & 2e-05 & 32 & 10\\\hline
\end{tabular}
}
\vspace*{-1mm}
\caption{The optimal hyperparameter settings when fine-tuning the RoBERTa base model on the SemEval 2018 Task 1 dataset for each emotion.}  
 \label{tab:hyperparameter_tune}
\end{table*}

\subsection{UED Metric Results}
\label{appendix:ml_results}
\blue{In Figure \ref{fig:ml_val_results} we show the results for rise rate and recovery rate for valence using the fine-tuned ML model on the PoKi dataset. Both rise rate and recovery rate increase with age. These trends support those seen when using the lexicon approach.
}

\blue{
In Figure \ref{fig:ml_ued} we show the UED metrics across grade for the discrete emotions (e.g., anger, fear, joy, and sadness). The trends for fear and sadness are similar to trends found when using the lexicon approach: emotion intensity, variability, rise rate and recovery rate increase across grades. 
Overall, the patterns of emotion change for anger are flatter across metrics. Perhaps anger is 
a more challenging emotion for automatic systems to detect \cite{SemEval2018Task1}. 
The average intensity for joy has a similar pattern to that of valence. These two emotions could appear similar to the ML model resulting in similar trajectories.  }








\begin{figure*}[h]
\vspace*{-6mm}
    \centering
    \begin{subfigure}[b]{0.49\textwidth}
    \centering
    \includegraphics[width=\textwidth]
    {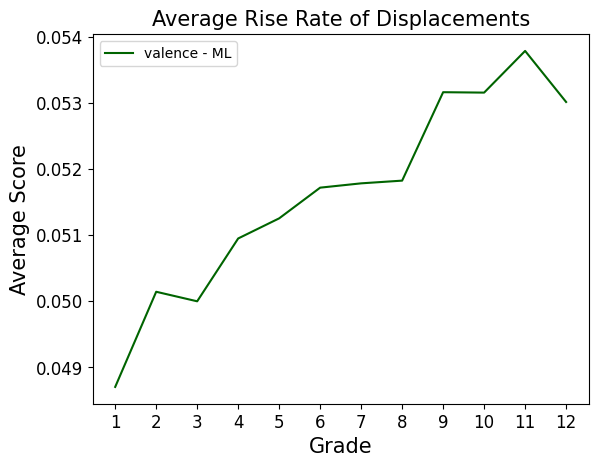}
    \label{fig:rise_rate_ml}
    \end{subfigure}
    \hfill
    \begin{subfigure}[b]{0.49\textwidth}
    \centering
    \includegraphics[width=\textwidth]
    {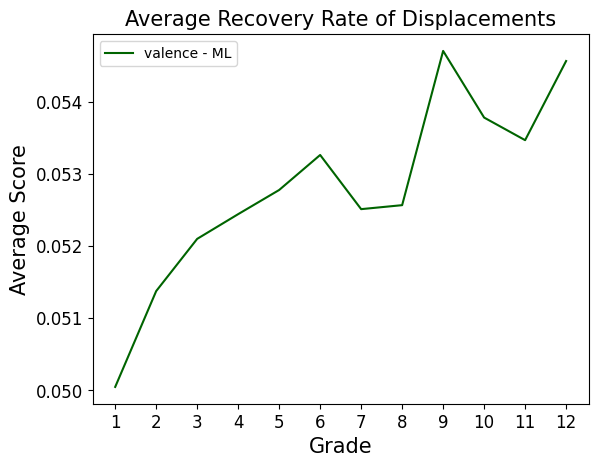}
    \label{fig:recovery_rate_ml}
    \end{subfigure}
     \vspace*{-3mm}
    \caption{Rise rate and recovery rate for valence using the ML \textit{n-gram} approach on the PoKi dataset.
    }
    \label{fig:ml_val_results}
    \vspace*{-3mm}
\end{figure*}

\begin{figure*}[h]
    \centering
    \begin{subfigure}[b]{0.49\textwidth}
    \centering
    \includegraphics[width=\textwidth]{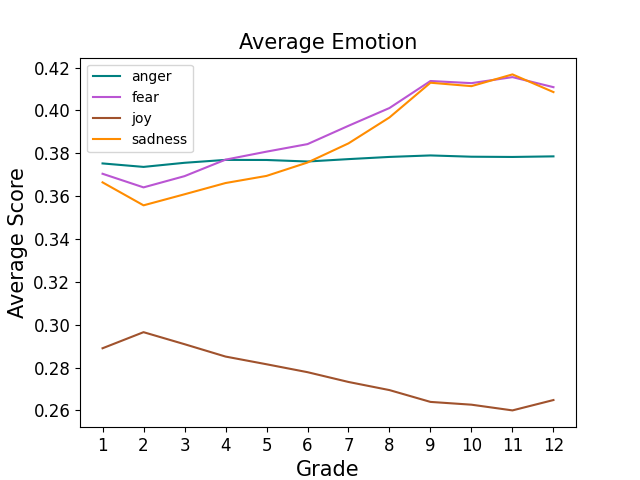}
    \label{fig:emo_mean_ml_discrete}
    \end{subfigure}
    \hfill
    \begin{subfigure}[b]{0.49\textwidth}
    \centering
    \includegraphics[width=\textwidth]
    {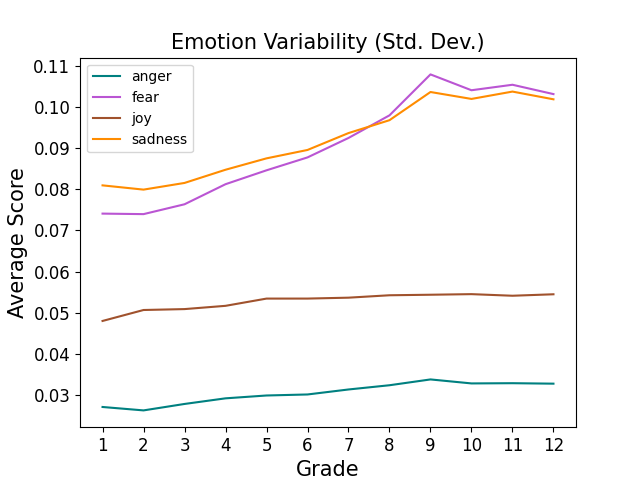}
    \label{fig:emo_std_ml_discrete}
    \end{subfigure}
    \vskip\baselineskip
    \centering
    \begin{subfigure}[b]{0.49\textwidth}
    \centering
    \includegraphics[width=\textwidth]
    {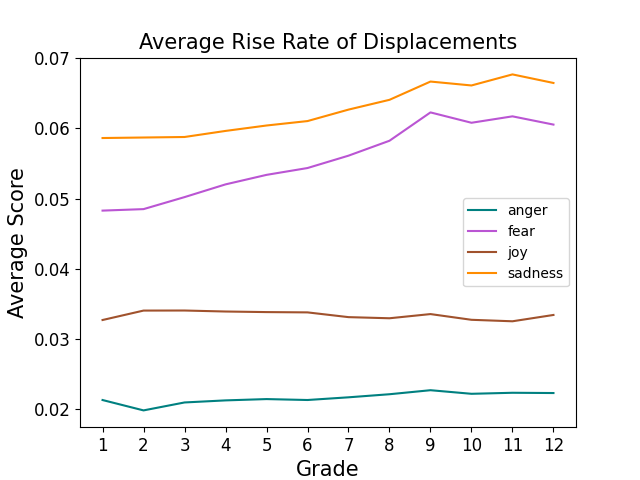}
    \label{fig:rise_rate_ml_discrete}
    \end{subfigure}
    \hfill
    \begin{subfigure}[b]{0.49\textwidth}
    \centering
    \includegraphics[width=\textwidth]
{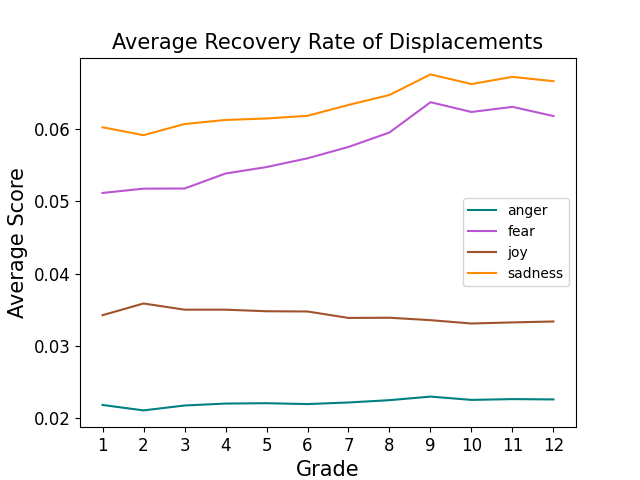}
    \label{fig:recovery_rate_ml_discrete}
    \end{subfigure}
    \caption{UED metrics for anger, fear, joy and sadness using the ML \textit{n-gram} approach on the PoKi dataset. 
    }
    \label{fig:ml_ued}
\end{figure*}

\end{document}